\documentclass[twoside,11pt]{article}

\usepackage{jmlr2e}
\usepackage[german,english]{babel}

\providecommand{\tabularnewline}{\\}

\ShortHeadings{Truecluster}{Oehlschl\"agel}
\firstpageno{1}

\begin{document}

\title{Truecluster: robust scalable clustering\\
with model selection}

\author{\name Jens Oehlschl\"agel \email jens.oehlschlaegel@truecluster.com}

\editor{}

\maketitle

\begin{abstract}
Data-based classification is fundamental to most branches of science. While recent years have brought enormous progress in various areas of statistical computing and clustering, some general challenges in clustering remain: model selection, robustness, and scalability to large datasets. We consider the important problem of deciding on the optimal number of clusters, given an arbitrary definition of space and clusteriness. We show how to construct a cluster information criterion that allows objective model selection. Differing from other approaches, our truecluster method does not require specific assumptions about underlying distributions, dissimilarity definitions or cluster models. Truecluster puts arbitrary clustering algorithms into a generic unified (sampling-based) statistical framework. It is scalable to big datasets and provides robust cluster assignments and case-wise diagnostics. Truecluster will make clustering more objective, allows for automation, and will save time and costs. Free R software is available.
\end{abstract}


\section{Introduction}

The power of modern computers has revolutionized the way we do statistical analysis. Computer-intensive simulation methods, such as calculation of standard errors via bootstrapping in frequentist statistics \citep{Efron:1979,EfronTibshirani:1993} or MCMC methods in Bayesian statistics, have become increasingly important in order to address the two big themes in statistical data mining: `prediction' and `clustering'. An important problem in both areas is model selection: how to find a statistical model optimally adapted to the true patterns in a data population, while avoiding overfit to the sample?

In predictive modeling, early approaches tried to penalize too much flexibility by subtracting the parametric model's degrees of freedom from the log-likelihood, namely $AIC$ \citep{Akaike:1973,Akaike:1974} and $BIC$ \citep{Schwarz:1978}. Other approaches made use of cross-validation or bootstrapping with shrinking in order to estimate and correct the amount of overfit, cf. \citet{Harrell:2001}. Recent scalable methods combine resampling and model averaging: they optimize model complexity, minimize overfit, and result in very robust predictions, for example, bagging \citep{Breiman:1996} or random forests \citep{Breiman:2001}. 

Contrary to these advances in predictive modeling, cluster analysis has not yet reached the same maturity in scalability, robustness, and model selection: while a multitude of powerful algorithms has been developed \citep{HalkidiBatistakisVazirgiannis:2001, berkhin:2002, ZaianeFossLeeWang:2002, JainTopchyLawBuhmann:2004}, the field lacks a coherent statistical framework for model selection: up to now the decision for an optimal number of clusters within a single model class has not been generically solved, not to mention comparisons across model classes. Identifying the correct number of clusters is of great practical importance and---as a general problem---has not been considered solved for more than 50 years (\citeauthor{Thorndike:1953}, \citeyear{Thorndike:1953}; \citeauthor{Everitt:1979}, \citeyear{Everitt:1979}; \citeauthor{Gordon:1999}, \citeyear{Gordon:1999}, chapt.~3.5; \citeauthor{EverittLandauLeese:2001}, \citeyear{EverittLandauLeese:2001}, chapt.~5.5; \citeauthor{GordonVichi:2001}, \citeyear{GordonVichi:2001}; \citeauthor{Dolnicar:2003}, \citeyear{Dolnicar:2003}). 

We present here a generic statistical framework for the selection of the optimal number of clusters within a single model class. We finally derived the framework under three major restrictions: the method must be scalable to large datasets, must deliver robust cluster results, and must provide useful case-wise diagnostics. Furthermore, we were looking for an approach that can easily be implemented without proprietary software, preferably in R \citep{R:W1061} in order to make it available for research and application \citep{Milligan:1996}. We organize the paper as follows: in Section \ref{sec:problem}, we define the problem and review related work, in Section \ref{sec:truecluster},  we introduce the truecluster framework, in Section \ref{sec:MMCC}, we briefly summarize the truecluster matching and voting scheme, in Section \ref{sec:CIC}, we introduce the \emph{cluster information criterion} $CIC$ and then discuss scalability in Section \ref{sec:scalability}. In Section \ref{sec:example}, we illustrate the truecluster method applied to a real world example and, finally, in Section \ref{sec:discussion}, we discuss the benefits and restrictions of truecluster. Appendix A shows truecluster results for some illustrative artificial datasets. Truecluster software is work-in-progress \citep{R:truecluster} and currently offers truecluster matching (see \texttt{?matchindex}) and calculation of the cluster information criterion (see \texttt{?CIC}). 

\section{Problem definition and related work\label{sec:problem}}

We consider the following general setup for cluster analysis: $N$ cases in an $M$-dimensional feature space sampled from an infinite population will be classified into an optimal number of $K$ distinct clusters, given some definition of `clusteriness' represented by a base-cluster algorithm that takes $K$ as input parameter. The number of possibilities to assign cases to clusters grows exponentially with the number of cases. Checking all the possibilities for an optimal solution is prohibitive even with tomorrow's computers. This is one reason for the existence of so many cluster algorithms---and the need to validate the resulting models. 

Our problem definition excludes base cluster algorithms that do not allow the number of clusters to be specified in advance, for example, the k-th nearest neighbor selection method by \citet{WongSchaack:1982}. Such algorithms deliver an automatically emerging number of clusters, but usually require subjective choice of other---often continuous---input parameters that are even more difficult to optimize. For obvious reasons, we also exclude well-defined decison problems (and algorithms), where the optimal number of clusters follows from minimizing a loss-function, see \citet[chapter 28, model comparison, and Occam's razor]{MacKay:2003}. For example, the affinity propagation algorithm of \citet{FreyDueck:2007}, which automatically delivers an optimal set of 'exemplars', is in fact a method for cost optimization: both inputs, objective `similarities' and subjective `preferences', can be interpreted as (inverse) costs.

We briefly summarize some of the existing approaches to cluster validation and model selection and point out important restrictions for each of them. We do not consider methods here that are designed to only work with one specific cluster algorithm such as the maximum spanning tree stopping rule for single linkage agglomeration \citep{Krolak-SchwerdtEckes:1992}. 

Approaches that calculate a goodness-of-fit (GOF) criterion with respect to a definition of clusteriness, often termed `internal criteria' \citep{Milligan:1981}, are the most widespread. An example is the silhouette width \citep{Rousseeuw:1987} which evaluates the quality of separation of convex partitions (with respect to case dissimilarity). The 30 cluster evaluation indices compared by \citet{MilliganCooper:1985} and the 15 indices compared by \citet{DimitriadouDolnicarWeingessel:2002} are based on the GOF approach: cluster solutions are evaluated in the $N$ x $M$ feature space or in the $N$ x $N$ dissimilarity matrix. Evaluating GOF has its justification in the fact that most cluster algorithms do not guarantee finding an optimal clustering and are often sensitive to outliers, for example, the widespread k-means \citep{KMEANS_MacQueen:1967}. Since each GOF index relates to a specific definition of clusteriness, `no single superior procedure can be recommended' \citep{Dolnicar:2003} as a general method for deciding on the optimal number of clusters. Furthermore, using GOF as a decision method is complicated by the fact that GOF is often biased with the number of clusters and decision makers are asked to identify a `knee' \citep[p. 129]{HalkidiBatistakisVazirgiannis:2001} in a plot of GOF versus the number of clusters. In order to work around such bias, some GOF indices such as the cubic clustering criterion \citep{Sarle:1983} or the gap statistic \citep{TibshiraniWaltherHastie:2001} relate GOF to the expected GOF under a null hypothesis, see below. 

More systematic approaches to decisions about the optimal number of clusters have been developed in the context of parametric probabilistic models. \citet{Smyth:1996} distinguished three approaches: hypothesis testing \citep{Bock:1996}, full Bayesian analysis \citep[part IV]{MacKay:2003} such as AutoClass \citep{CheesemanStutz:1996}, and penalized likelihood such as $BIC$ \citep{MCLUST_BanfieldRaftery:1993,KassRaftery:1995,FraleyRaftery:1998}. Smyth concludes that \emph{`In theory, the full Bayesian approach is fully optimal and probably the most useful of the three methods listed above. However, in practice it is cumbersome to implement, it is not necessarily straightforward to extend to non-Gaussian problems with dependent samples, and the results will be dependent in a non-transparent manner, on the quality of the underlying approximations and simulations. Thus, there is certainly room for exploring alternative models.'} \citep[p. 127]{Smyth:1996}. For example, the EM-algorithm underlying the above mentioned $BIC$ does not scale easily to big samples. Smyth suggests Monte Carlo cross validation which he found performed as well as AutoClass and better than the $BIC$. \citet{ChickeringHeckerman:1997} found the $BIC$ to work reasonably for model selection and confirmed superiority of AutoClass over $BIC$ for model averaging. \citet{DimitriadouDolnicarWeingessel:2002} also reported problems with the $BIC$ in latent class analysis. In summary, for parametric probabilistic models, acceptable model selection methods are available. Still, correctness of these methods rely on correct parametric assumptions and the methods don't easily translate to non-parametric definitions of clusteriness. 

The most general approach to cluster validation is given when feature space or dissimilarity space are completely ignored and we just compare agreement of cluster results from several `disturbed' solutions, for example, solutions from bootstrapping, cross-validation or feature sampling. Transferring ideas from prediction model validation, several authors have stressed the importance of validating independent (non-overlapping) sub-samples in order to avoid bias \citep{DudoitFridlyand:2002,TibshiraniWaltherBotsteinBrown:2001}. However, even independent resampling schemes do not achieve true independence with respect to clustering: spatial neighbors have a higher likelihood of being clustered together. Therefore, even `random-corrected' agreement indices, such as Cohen's kappa \citep{Cohen:1960} or the random-corrected version \citep[crand]{HubertArabie:1985} of the rand index \citep{Rand:1971}, will show `non-random' agreement in non-clustered random data. In order to work around such bias, \citet{DudoitFridlyand:2002} have suggested relating cross-validation agreement to performance of the same agreement index (and the same base cluster algorithm) in simulating from a reference null distribution. 

We have seen that the difficulties with the GOF approach and cross-validated agreement indices have led to suggest resorting to assuming (or simulating from) a reference null distribution: the null hypothesis of random clustering. Such a null distribution is supposed to represent a neutral no-cluster situation; however, the choice of a null distribution is subjective and can influence the results. Take a 2-dimensional uniform random distribution: the square shape will induce artificial agreement for a 4-cluster k-means solution, something that will be different using a multivariate normal null distribution. Therefore, we are looking for an approach that treats the base cluster algorithm as a black box, works with any definition of clusteriness, and requires no assumptions about null-distributions, variable space or dissimilarity definitions.

\section{Truecluster\label{sec:truecluster}}

The first idea for truecluster dates back to 1996 when we tried to select the best number of clusters for a given base-cluster algorithm by evaluating the stability of each K-cluster solution via repeated split-half cross-validation. In 1997 we got to know the draft of Harrell's book and S software library \citep{Harrell:2001} that suggests bootstrapping for the validation and calibration of regression models. We were fascinated by the idea of comparing the stability of cluster models fitted to resamples of the same sizes as the original sample size, since in split-half samples stability could be biased downward. On the other hand, Harrell's approach involved comparisons between models built on overlapping data and, thus, could estimate stability biased upward. We experimented intensively with both approaches and came to the conclusion that both are biased. Surprisingly, we found a second source for a reduced split-half stability, due to outliers that---systematically--- are in one but not in the other split-half sample. Then we learned about Breiman's work on bagging \citep{Breiman:1996} and experimented with aggregating $K$ x $K$ agreement counts (or agreement statistics based on them) between many pairs of cluster solutions. In order to create $K$ x $K$ cluster agreement counts we would either need overlapping samples as in bootstrapping or---in the case of split-half---to assign the out-of-resample cases to the clusters found in resampling. We concluded that the $K$ x $K$ matrices don't contain enough information for proper model selection. Consequently, we turned to aggregating the cluster assigments themselves in a $N$ x $K$ matrix in order to create a cluster version of bagging, very similar---but not equal---to the \emph{BagClust1} algorithm by \citet{DudoitFridlyand:2003}. An aggregated $N$ x $K$ matrix would not only promise to contain enough information for model selection but also allow the creation of new, more robust cluster assignments with case-wise diagnostics and to scale expensive base cluster algorithms to larger samples. We decided not to go for aggregation of the even more informative $N$ x $N$ co-occurence counts as suggested by \citet[BagClust2]{DudoitFridlyand:2003} because this requires $\mathbf{O}(N^{2})$ space complexity. Given this decision, the truecluster approach involves two steps described in the next two sections: 

\begin{enumerate}
\item For each number of clusters $K$, aggregate the results of many resamples in an $N$ x $K$ matrix $\mathbf{C}_{K}$ using a specific matching and voting logic called \emph{multiple match cluster counts} $MMCC$ and convert to an $N$ x $K$ matrix $\hat{\mathbf{P}}_{K}$ that contains estimated probabilities $\hat{P}_{i,k}$: how likely it is for each case $i\in\{1...N\}$ to be assigned to cluster $k\in\{1...K\}$ across many resamples (see Section \ref{sec:MMCC}).
\item For each number of clusters $K$, calculate a \emph{cluster information criterion} $CIC_{K}$ using $\hat{\mathbf{P}}_{K}$ as input. Choosing the model with the highest $CIC_{K}$ gives the optimal number of clusters $K$ (see Section \ref{sec:CIC}).
\end{enumerate}

For simplicity, we drop the index $K$ from all the following notation.

\section{Multiple match cluster counts (MMCC)\label{sec:MMCC}}

\citet{DudoitFridlyand:2003} initialize \emph{BagClust1} by applying the base cluster algorithm to the complete sample and use this as a reference for permutating the labels of the bootstrap clusterings that are to be aggregated in $\mathbf{C}$. Because some base cluster algorithms don't scale to arbitrarily large samples this choice is not generally applicable. Aggregating votes from subsampling can help, but one needs sufficient overlap for permutation of the subsample cluster labels. Complete overlap can be achieved by assigning the out-of-resample cases to the resample clusters. Such prediction is often computationally less expensive compared to the base cluster algorithm. The resulting partially-predicted full-sample solutions can be used to initialize $\mathbf{C}$ and for the subsequent voting. Unlike \emph{BagClust1}, no single solution serves as a reference for label permutation, instead $\mathbf{C}$ itself is used, because $\mathbf{C}$ becomes a better cluster representation with ongoing voting, similar to the suggestion by \citet{DimitriadouWeingesselHornik:2002}. $\mathbf{C}$ (and $\hat{\mathbf{P}}$) can be interpreted as a fuzzy cluster solution and can aggregate results from fuzzy base cluster algorithms, however, both (fuzzy resample solutions and the $\mathbf{C}$ reference) need to be converted to a crisp clustering before doing the label permutation. This differs from fuzzy consensus clustering \citep{GordonVichi:2001, DimitriadouWeingesselHornik:2002} and guarantees that $\hat{\mathbf{P}}$ can be given a probability interpretation which is crucial for the $CIC$ evaluation. 

While consensus cluster methods for crisp \citep{StrehlGhosh:2002} or fuzzy cluster ensembles \citep{GordonVichi:2001, DimitriadouWeingesselHornik:2002} aim to minimize euclidean distances between cluster representations in the ensemble, the label permutation in $MMCC$ uses a different matching criterion. The reason for the difference is that consensus clustering tries to find the deterministic optimal representation of a finite cluster ensemble, whereas $MMCC$ is a probabilistic algorithm trying to converge an appropriate representation $\mathbf{C}$ for a single cluster model. Details of truecluster matching are given in a separate paper \citep{Oehlschlaegel:2007b} and free software is available \citep{R:truecluster}. The standard $MMCC$ algorithm can now be described as follows:

\begin{enumerate}
\item Create a $N$ x $K$ matrix $\mathbf{C}$ and initialize each cell $C_{i,k}$
with zero. 
\item Take a resample (with replacement) of size $n$, use a \emph{base cluster algorithm} to fit the $K$-cluster model $\mathbf{\mathbf{c}^{*}}$ to the resample. Then use a suitable \emph{prediction method} to determine cluster membership of the out-of-resample cases to get a complete cluster vector \textbf{$\mathbf{c^{'}}$} with $N$ elements $c_{i}^{'}$ .
\item For each row in $\mathbf{C}$, add one vote (add 1) to the column corresponding to the cluster membership in \textbf{$\mathbf{c^{'}}$}. 
\item Repeat step 2.
\item Estimate cluster memberships $\mathbf{\hat{c}}$ by a row-wise majority count in $\mathbf{C}$ (breaking ties at random), use the truematch \emph{algorithm} or \emph{heuristic} \citep{Oehlschlaegel:2007b} to align \textbf{$\mathbf{c^{'}}$} with $\hat{\mathbf{c}}$, and rename the clusters in \textbf{$\mathbf{c^{'}}$} like the corresponding clusters in $\hat{\mathbf{c}}$ . 
\item For each row in $\mathbf{C}$, add one vote (add 1) to the column corresponding to the cluster membership in \textbf{$\mathbf{c^{'}}$}. 
\item Repeat from step 4 until some reasonable \emph{convergence criterion} is reached. 
\item Divide each cell in $\mathbf{C}$ by its row-sum to get a matrix of estimated cluster membership probabilities $\hat{\mathbf{P}}$ .
\end{enumerate}

Remark 1: Resampling with replacement was chosen because this can be applied to samples of any size and reflects the usual assumption that the sample stems from an infinite population. With this choice, special care is needed to avoid difficulties with duplicate cases, for example, duplicated initial centers with the k-means base algorithm. A finite-population setting, a very large sample size or a base cluster algorithm's intolerance to duplicated values might justify a different sampling scheme.

Remark 2: When a base cluster algorithm and a prediction method are computationally expensive and scaled to very large samples, a variation of $MMCC$ might scale better: subdivide the sample into sufficiently overlapping subsamples and integrate these to get an initial $\mathbf{C}$, similar to suggestions by \citet{StrehlGhosh:2002}. Then match the subsample solutions without prediction and vote only for the cases in the subsamples. Row-sums of $\mathbf{C}$ will no longer be equal. 

Remark 3: Estimation of $\hat{\mathbf{P}}_{K}$ is robust as a consequence of the resample aggregation. Like in bagging, the influence of outliers is reduced because they are not sampled into all resample models. Unlike deterministic optimization procedures that are exposed to outlier influence in each convergence step, outliers can only influence some steps during the stochastic convergence of the $\hat{\mathbf{P}}_{K}$ matrix. Resample aggregation can be interpreted as a stochastic version of the EM-algorithm \citep{DempsterLairdRubin:1977}: estimating the missing class labels $\hat{\mathbf{c}}$ from \textbf{$\mathbf{C}$} is clearly an e-step. Matching the next resample solution \textbf{$c^{'}$} to the current best estimate $\hat{\mathbf{c}}$ is a maximization step. Unlike classical EM, this is not an m-step maximizing the full model but only an optimized voting improving the model stochastically. Details on convergence will be provided elsewhere.

Remark 4: For the variation in cluster solutions, we focused on case sampling since this gives rise to clear statistical interpretation of the probabilitites in $\hat{\mathbf{P}}_{K}$. However, analogous to random forests \citep{Breiman:2001}, it should be possible to extend truecluster to attribute sampling, cf. \citet{StrehlGhosh:2002}. It is obvious that whatever sampling scheme is used, it must be equal across all $K$ in order to obtain comparable $CIC_{K}$. 

Remark 5: $MMCC$ might be seen as a special case of creating and aggregating a cluster ensemble. Because truecluster focuses on identifying the best number of clusters $K$ rather than creating  consensus across different $K$ or across different base cluster algorithms, ensembles can be very large and there are as many ensembles as candidates for the best $K$ which can easily exhaust computer memory. For example, the methods suggested by \citet{StrehlGhosh:2002} would require holding all cluster solutions of all resamples simultaneously in the memory. Implications for software architecture will be provided elsewhere.

All the calculations in this paper have used resampling with replacement, prediction, and the truematch \emph{heuristic} that is obtained with \texttt{matchindex(method="heuristic")} \citep{R:truecluster}.

\section{Cluster information criterion (CIC)\label{sec:CIC}}

After having introduced the truecluster framework, this paper focuses on the evaluation of the results of the $MMCC$ aggregated voting: how to condense the information in $\hat{\mathbf{P}}$ to a single value, guiding selection of optimal $K$: the \emph{cluster information criterion ($CIC$)}. 

Predictive class modeling can be described as declaring the existence of $K$ classes with i.i.d. cases and it is clear that more classes result in more homogeneous distributions and higher log-likelihood. To cite Gideon Schwarz: ``In such cases the maximum likelihood principle invariably leads to choosing the highest possible dimension. Therefore it cannot be the right formalization of the intuitive notion of choosing the `right' dimension'' (1978, p. 461). Thus, Akaike's $AIC$ and Schwarz' $BIC$ penalize model certainty (the log-likelihood) by model complexity (the degrees of freedom). For the resample aggregation matrix $\hat{\mathbf{P}}$, the contrary is true: more complex models are usually---with some exceptions---less stable. In this case, the maximum certainty principle invariably leads to choosing the lowest possible number of clusters. Therefore, it cannot be the right formalization of the intuitive notion of choosing the right number of clusters. It is important to realize that this limitation of certainty-only approaches extends to all stability-only approaches, including those that look for non-random stability. Following the logic to correct the shortcoming of a certainty-only approach, we suggest \emph{rewarding} the model certainty for the model complexity. Therefore, we define the \emph{cluster information criterion} of the $K$-cluster model as 

\begin{equation}
\begin{array}{ccccc}
CIC & = & model\, certainty & + & model\, information\\
 & = & model\, information & - & model\, uncertainty\end{array}\label{eq:CIC}\end{equation}

a trade-off between model information and model uncertainty, both measured in bits \citep{Shannon:1948}. The $CIC$ trades off information against uncertainty in evaluating the combination of the base cluster algorithm and prediction method. Given a fixed base cluster algorithm, prediction method, and resample size, the $CIC$ can be used for objective automatic model selection---without the need to specify a null reference distribution. In the following section, we derive the calculation of model information, model uncertainty, and further diagnostics.

Imagine a system with $K$ states and let's begin with the simplifying assumption that all states are equally likely. Let's assume we don't know the system's actual state and call this our \emph{uncertainty} and let's measure this in bits

\[uncertainty_{K}=log_{2}K\]

Let's define \emph{information} as the reduction of uncertainty when we get to know the actual state $k\in\{1...K\}$

\[information_{K}=log_{2}K\]

So for a system with $K=4$ states, our uncertainty is $log_{2}(4)=2$ bit and we can gain 2 bit of information. Now let's generalize and introduce different probabilities $p_{k}$ for our states. It is obvious that we don't have any uncertainty if $p=1$ for one state and $p=0$ for the other states. Uncertainty is maximal if all states have equal probability $p=1/K$ . The amount of uncertainty (or gainable information) of such a probabilistic system can be quantified as its \emph{entropy}

\begin{equation}entropy=-\sum_{k=1}^{K}p_{k}log_{2}p_{k}\label{eq:H}\end{equation}

It is instructive to note that for equal probabilities $p=1/K$ this simplifies as it should to

\[entropy_{K}=-\sum_{k=1}^{K}\frac{1}{K}log_{2}\frac{1}{K}=log_{2}K\]

Now we can give our probabilistic system the interpretation of a \emph{random distribution} and recognize that the entropy is the weighted average of $-log_{2}p_{k}$ (weighted by state probability). In other words: entropy is the expected value of the information gained after randomly sampling one observation from our distribution. Equation \ref{eq:H} can be used to measure the model information of a cluster model declaring $K$ clusters of a certain size and using crisp assignments of cases to clusters. 

The classical measure of model certainty in predictive modeling is the log-likelihood, representing the probability of the data observed given the model. Applying this logic to our $\hat{\mathbf{P}}$ matrix we get

\begin{equation}pseudo~log_{2}likelihood=\sum_{i=1}^{N}log_{2}\hat{P}_{i,c}\label{eq:LL}\end{equation}

where $\hat{P}_{i,c}$ denotes for each case $i$ the probability of the most frequently voted cluster. Two things are wrong with Equation \ref{eq:LL}. First, it's not really a likelihood because the observations are not independent with respect to their cluster assignments; second, no crisp cluster memberships $c$ have been observed. Instead, our model $\mathbf{\hat{P}}$ states probabilities $\hat{P}_{i,\cdot}$ estimating how likely it is that case $i$ belongs to each of the clusters. Therefore, we generalize Equation \ref{eq:LL} to the non-crisp case. Following Equation \ref{eq:H} we replace per case the value of $log_{2}\hat{P}_{i,c}$ in Equation \ref{eq:LL} through the expected value $\sum_{k=1}^{K}\hat{P}_{i,k}log_{2}\hat{P}_{i,k}$  (or zero if $\hat{P}_{i,c}=1$) across all clusters

\begin{equation}model~uncertainty=-\frac{1}{N}\sum_{i=1}^{N}\sum_{k=1}^{K}\hat{P}_{i,k}log_{2}\hat{P}_{i,k}\label{eq:Uncertainty}\end{equation}

In this definition of \emph{model uncertainty}, we have switched the sign and additionally divided it by the sample size $N$ in order to make our measure independent of sample size. For a crisp matrix with all cluster member probabilities $\hat{P}_{i,c}=1$ and all other $\hat{P}_{i,\cdot}=0$, Equation \ref{eq:Uncertainty} reduces to Equation \ref{eq:LL}, but generally our model uncertainty evaluates all cells of $\mathbf{\hat{P}}$. Equation \ref{eq:Uncertainty} can be interpreted as a conditional entropy \citep{MacKay:2003} of clusters, given the cases. In the context of fuzzy clustering, Equation \ref{eq:Uncertainty} is known as partition entropy \citep{BezdekWindhamEhrlich:1980}, which---without further correction---is known to depend on the number of clusters. 

Now being equipped with a definition of model uncertainty, we can easily show---following Schwarz---that the uncertainty alone is not sufficient to evaluate cluster models. The following three example matrices $\hat{\mathbf{P}}$ all have the same uncertainty (and log-likelihood)---zero---but they obviously differ in \emph{model information}:

\[
\begin{array}{cc}
1 & 0\\
1 & 0\\
1 & 0\\
1 & 0\\
0 & 1\\
0 & 1\\
0 & 1\\
0 & 1\\
0 & 1\\
0 & 1\\
0 & 1\\
0 & 1\\
0 & 1\\
0 & 1\\
0 & 1\\
0 & 1\end{array}\neq\begin{array}{cc}
1 & 0\\
1 & 0\\
1 & 0\\
1 & 0\\
1 & 0\\
1 & 0\\
1 & 0\\
1 & 0\\
0 & 1\\
0 & 1\\
0 & 1\\
0 & 1\\
0 & 1\\
0 & 1\\
0 & 1\\
0 & 1\end{array}\neq\begin{array}{cccc}
1 & 0 & 0 & 0\\
1 & 0 & 0 & 0\\
1 & 0 & 0 & 0\\
1 & 0 & 0 & 0\\
0 & 1 & 0 & 0\\
0 & 1 & 0 & 0\\
0 & 1 & 0 & 0\\
0 & 1 & 0 & 0\\
0 & 0 & 1 & 0\\
0 & 0 & 1 & 0\\
0 & 0 & 1 & 0\\
0 & 0 & 1 & 0\\
0 & 0 & 0 & 1\\
0 & 0 & 0 & 1\\
0 & 0 & 0 & 1\\
0 & 0 & 0 & 1\end{array}\]

The first matrix has two classes with probabilities $\left\{ \frac{1}{4},\frac{3}{4}\right\} $ and, thus, $entropy=-\frac{1}{4}log_{2}\frac{1}{4}-\frac{3}{4}log_{2}\frac{3}{4}\simeq0.8$,
the second has $entropy=-\frac{1}{2}log_{2}\frac{1}{2}-\frac{1}{2}log_{2}\frac{1}{2}=1$ and, the third has $entropy=-4\frac{1}{4}log_{2}\frac{1}{4}=2$. Clearly the amount of information delivered by these models is different---the last one being the most informative. The reader will have noted that this was a crisp example. Analogous to the model uncertainty, we now generalize from the information of a crisp cluster membership vector to the non-crisp case: we want to avoid information loss resulting from considering marginals only. The marginal cluster probabilities $p_{k}$ in Equation \ref{eq:H} can be interpreted as the column-means of (a crisp) $\mathbf{\hat{P}}$. This suggests actually defining $\hat{p}{}_{k}$ as the column-means of the non-crisp $\mathbf{\hat{P}}$ and to focus on the amount of information gained on average by knowing $\hat{P}_{i,\cdot}$ when randomly sampling one case: we define $\mathbf{\hat{D}}$ as the conditional information of cases, given the clusters, 

\begin{eqnarray}
\hat{p}_{k} & = & \frac{\sum_{i=1}^{N}\hat{P}_{i,k}}{\sum_{k=1}^{K}\sum_{i=1}^{N}\hat{P}{}_{i,k}}\label{eq:ClusterProbability}\\
\hat{D}_{i,k} & = & -\hat{P}_{i,k}log_{2}(1-|\hat{P}_{i,k}-\hat{p}_{k}|)\label{eq:WeightedLogDeviation}
\end{eqnarray}

which evaluates the difference between the case probabilities $\hat{P}_{i,\cdot}$ and the average column probabilities $\hat{p}_{k}$. It can be easily seen that for crisp cluster assignments Equation \ref{eq:WeightedLogDeviation} reduces to Equation \ref{eq:H}. 

Measures such as Equations \ref{eq:H} and \ref{eq:WeightedLogDeviation} signal too much information for over-complex models. For a model that assigns each case into its own private cluster, we formally get $entropy=log_{2}N$, where in fact the model does not deliver any information at all. Therefore, we need to penalize $\hat{D}_{i,k}$ for model complexity. Simply penalizing for the degrees of freedom $K$ by scaling down with $(1-\frac{K}{N})$ implicitly assumes equal cluster sizes. As a more general measure for model complexity than $\frac{K}{N}$, we suggest the \emph{relative model complexity} $RMC$

\begin{equation}
RMC=\frac{2^{\left(\sum_{k=1}^{K}\hat{p}_{k}log_{2}\hat{p}_{k}\right)}-1}{N-1}\label{eq:RelativeModelComplexity}
\end{equation}

which takes values between 0 (no model complexity) and 1 (maximum model complexity). We now can define the cell-wise model information $\hat{I}_{i,k}$ by penalizing $\hat{D}_{i,k}$ for $RMC$ (Equation \ref{eq:CellInformation}). By summing rows and averaging over columns of \textbf{$\hat{\mathbf{I}}$}, we obtain the model information (Equation \ref{eq:Information}) that we need in order to estimate the $CIC$ in Equation \ref{eq:CIC}. 

\begin{eqnarray}
\hat{I}_{i,k} & = & \hat{D}_{i,k}\cdot(1-RMC)\label{eq:CellInformation}\\
model\, information & = & \frac{1}{N}\cdot\sum_{k=1}^{K}\sum_{i=1}^{N}\hat{I}_{i,k}\label{eq:Information}
\end{eqnarray}

For diagnostic purposes, we might want to express the $CIC$ as a function of cell-wise components $CIC_{i,k}$ 

\begin{equation}
\begin{array}{ccc}
CIC & = & \frac{1}{N}\cdot\sum_{i=1}^{N}\sum_{k=1}^{K}CIC_{i,k}\\
 & where & \\
CIC_{i,k} & = & \hat{I}_{i,k}-(-\hat{P}_{i,k}log_{2}\hat{P}_{i,k})\\
 & = & \hat{P}_{i,k}\left(log_{2}\frac{\hat{P}_{i,k}}{(1-|\hat{P}_{i,k}-\hat{p}_{k}|)^{(1-RMC)}}\right)
\end{array}\label{eq:CellwiseCIC}
\end{equation}

The model information in Equation \ref{eq:Information} quantifies how much the model tells us, on average, about a case randomly drawn from the sample (when replacing the marginal $\hat{p}{}_{k}$ by the case-specific row $\hat{P}_{i,\cdot}$). The model uncertainty (Equation \ref{eq:Uncertainty}) quantifies how much uncertainty, on average, the case-specific model claims involve. The $CIC$ finally is the expected value (across cases) of the case-wise expected value of the amount of information delivered by a ratio that becomes big, if cluster $k$ has high probability in case $i$, without having high probability in general (and without beeing penalized for over-complexity). Analyzing $CIC_{i,k}$ over cases or clusters can give valuable diagnostic insights. Finally, as an easy-to-interpret case-wise diagnostic, we suggest the \emph{generalized silhouette diagnostic (GSD)} 

\begin{equation}
GSD_{i}=\frac{\hat{P}_{i,c}}{\hat{P}_{i,c}+\hat{P}_{i,c2}}\cdot2-1\label{eq:GSD}
\end{equation}

ranging from 0 (ambiguous assignment) to 1 (unambiguous assignment) where $\hat{P}_{i,c2}$ is the estimated probability for the second best cluster per case. Functions for calculating $CIC$ and $GSD$ are available in R package \texttt{truecluster} \citep{R:truecluster}, see \texttt{?CIC}.

\section{Scalability\label{sec:scalability}}

$CIC$ model comparison is a computationally intensive method. Comparing 10 cluster models with 1,000 resamples requires 10,000 applications of the base cluster algorithm. This appears to be expensive but this standardized method is cheaper than ad-hoc manual model comparison; more importantly, it is scalable to big samples: the critical scalability component of resample aggregation is the base cluster algorithm. If it is scalable, truecluster is scalable as well. If the base cluster algorithm scales badly (takes too long or too much computer memory to handle \emph{N} cases), truecluster still allows fitting the model to the full dataset if the following assumptions are met: the base cluster algorithm scales to \emph{n} cases, this size of resample is sufficient to catch the complexity of the true cluster pattern, and the prediction method scales sufficiently. In very large samples with $n\ll N$, the critical component of the resample aggregation is the prediction method which is needed to classify those cases not in the resample. If no specific scalable prediction method is available, we can always resort to 1st nearest neighbor prediction. Determining the nearest neighbor of each datapoint is needed only once and can then be used for each resample prediction and for each $K$. Naive nearest neighbor identification has time complexity $\mathbf{O}(N^{2})$. Unless dimensionality is too high, kd-trees \citep{Bentley:1975} can speed up nearest neighbor identification, especially when exploiting the `all nearest neighbor' situation \citep{GrayMoore:2000}. When following the $MMCC$ Remark 2, no prediction is needed and voting is done in batches which actually scales the base cluster algorithm close to $\mathbf{O}(N)$.
Thus, truecluster scales arbitrary base cluster algorithms to large datasets depending on the scalability of the base cluster algorithm with $\mathbf{O}(N)$ (or an upper bound of $\mathbf{O}(N^{2})$) for time complexity and space complexity $\mathbf{O}(N \cdot K)$ (for one cluster solution given K).
Truecluster computations can be accelerated by distributing subtasks on parallel computing nodes. When comparing several models, each $\hat{\mathbf{P}}_{K}$ can be fitted on a separate node. Furthermore, in fitting each $\hat{\mathbf{P}}_{K}$, computations for $r$ resamples can be distributed across $r$ separate nodes.

\section{Example\label{sec:example}}

Mahon \citep{CampbellMahon:1974} recorded data on 200 specimens of Leptograpsus variegatus crabs on the shore in Western Australia. This occurs in two colour forms, blue and orange, and he collected 50 of each form of each sex and made five physical measurements. These were carapace (shell) length CL and width CW, the size of the frontal lobe FL and rear width RW, and the body depth BD. The latter was measured somewhat differently for males and females. This dataset has frequently been re-analyzed \citep{VenablesRipley:1994} and is publicly available \citep{R:MASS}. 

\begin{figure}[ht]
\includegraphics[%
  bb=66bp 189bp 530bp 653bp,
  width=1.0\columnwidth,
  keepaspectratio,
  angle=-90]{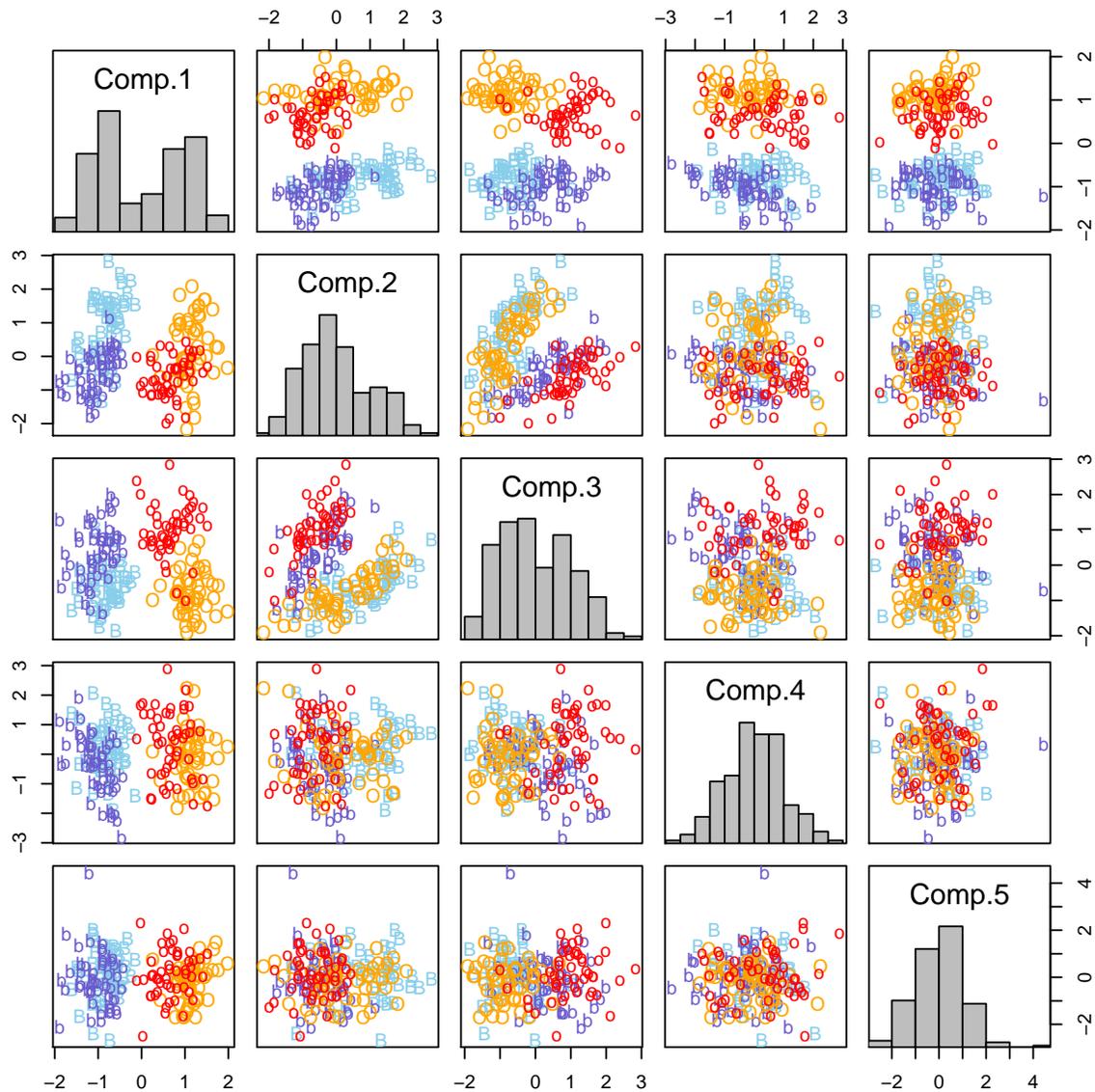}

\caption{Crab data in principal component space\protect \\
(B,b,O,o code blue and orange, males and females)\label{cap:Crabs-data-in-principal-component-space}}
\end{figure}

While the original analysis asks whether there are two morphologically distinct species or not, as an illustrative example we ask here whether a cluster algorithm will detect the four true classes in the data. We choose the measurement with the largest scale (CW) as an indicator of individual crab size and express the other four measurements by their ratio to CW, instead of their absolute size. CW itself is sufficiently symmetric so we do not transform CW to log scale. These measurements are then sphered using principal component analysis (using the correlation matrix) to define the cluster space. The data is shown as a scatterplot matrix in Figure \ref{cap:Crabs-data-in-principal-component-space}.

Looking for convex clusters, we choose partitioning around medoids (PAM) \citep{KaufmanRousseeuw:1990} as our base cluster algorithm and as a prediction method we assign out-of-resample cases to the closest medoid (in euclidean space). We begin the analysis by applying standard PAM for 2 to 10 clusters as implemented in R \citep{R:cluster}. Table \ref{cap:Evaluation-of-PAM-cluster-models} shows the arithmetic means of the GOF silhouette widths. Following this criterion, the 5-cluster solution appears to be best. The table also shows the information, uncertainty, and $CIC$ from the respective truecluster models based on 1000 bootstrap samples (n=200) or on 1000 subsamples of size n=100 (drawn with replacement). According to both truecluster models, a 5-cluster solution is rejected: the 2-cluster solution has the lowest uncertainty and the 4-cluster solution has---correctly---the best $CIC$. 

Table \ref{cap:Agreement-of-standard-and-truecluster-PAM} shows the agreement of the cluster solutions with the true classes: truecluster provides robust optimized solutions that show excellent agreement (and better agreement with true classes than standard PAM solutions). 

The truecluster bootstrap 200 and subsample 100 solutions disagree only in one case, the case with the lowest generalized silhouette diagnostic. GSDs of truecluster 200 and 100 correlate with r=0.986. Comparing standard PAM versus truecluster (200), we find that with respect to species, 143 cases are correctly classified by both cluster methods and in 27 cases both methods fail (Table \ref{cap:Failures-to-assign}). Of the remaining 30 cases, standard PAM fails in 29 and truecluster only in 1 case. Looking for disagreement with respect to species and gender, we find 22 cases, of which standard PAM fails in 17 and truecluster only in 5 cases.

\begin{table}[ht]\hfill
\begin{center}\begin{tabular}{|c||c||c|c|c|c|}
\hline 
&
standard&
\multicolumn{3}{c|}{truecluster 200}&
100\tabularnewline
\hline 
cluster\#&
silhouette&
information&
uncertainty&
CIC&
CIC\tabularnewline
\hline
\hline 
2&
0.131&
0.406&
0.736&
-0.330&
-0.404\tabularnewline
\hline 
3&
0.182&
0.859&
0.738&
0.121&
-0.267\tabularnewline
\hline 
4&
0.209&
1.012&
0.736&
0.199&
-0.119\tabularnewline
\hline 
5&
0.225&
1.055&
0.738&
0.108&
-0.196\tabularnewline
\hline 
6&
0.217&
1.042&
0.813&
-0.025&
{*}\tabularnewline
\hline 
7&
0.185&
1.058&
1.129&
-0.071&
{*}\tabularnewline
\hline 
8&
0.204&
1.050&
1.186&
-0.136&
-0.619\tabularnewline
\hline 
9&
0.214&
0.983&
1.351&
-0.368&
-0.811\tabularnewline
\hline 
10&
0.216&
0.981&
1.386&
-0.405&
{*}\tabularnewline
\hline
\end{tabular}\end{center}
\caption{Evaluation of PAM cluster models ({*} degenerated to fewer clusters) \label{cap:Evaluation-of-PAM-cluster-models}}
\end{table}

\begin{table}[ht]\hfill
\begin{center}\begin{tabular}{|c||c|c|c|c|c|c|}
\hline 
&
\multicolumn{3}{c|}{2 clusters x 2 species}&
\multicolumn{3}{c|}{4 clusters x 4 species/gender}\tabularnewline
\hline
&
standard&
100&
200&
standard&
100&
200\tabularnewline
\hline
\hline 
fraction matched{*}&
0.720&
0.920&
0.860&
0.845&
0.900&
0.905\tabularnewline
\hline 
kappa{*} \citep{Cohen:1960}&
0.440&
0.840&
0.720&
0.793&
0.867&
0.873\tabularnewline
\hline 
rand \citep{Rand:1971}&
0.595&
0.852&
0.758&
0.873&
0.909&
0.912\tabularnewline
\hline 
crand \citep{HubertArabie:1985}&
0.190&
0.704&
0.516&
0.663&
0.755&
0.765\tabularnewline
\hline
\end{tabular}\end{center}
\caption{Agreement of standard and truecluster PAM with true classes\protect \\
({*} after matching clusters)\label{cap:Agreement-of-standard-and-truecluster-PAM}}
\end{table}

\begin{table}[ht]\hfill
\begin{center}\begin{tabular}{|r|c||c|c|}
\hline 
&
code&
2 species&
4 species/gender\tabularnewline
\hline
\hline 
both OK&
o&
143&
164\tabularnewline
\hline 
standard PAM fails&
s&
29&
17\tabularnewline
\hline 
truecluster PAM fails&
t&
1&
5\tabularnewline
\hline 
both fail&
x&
27&
14\tabularnewline
\hline
\hline 
TOTAL&
&
200&
200\tabularnewline
\hline
\end{tabular}\end{center}

\caption{Failures to assign true classes\label{cap:Failures-to-assign}}

\end{table}

\begin{figure}[ht]
\begin{center}
\includegraphics[%
  width=8.6cm,
  angle=-90]{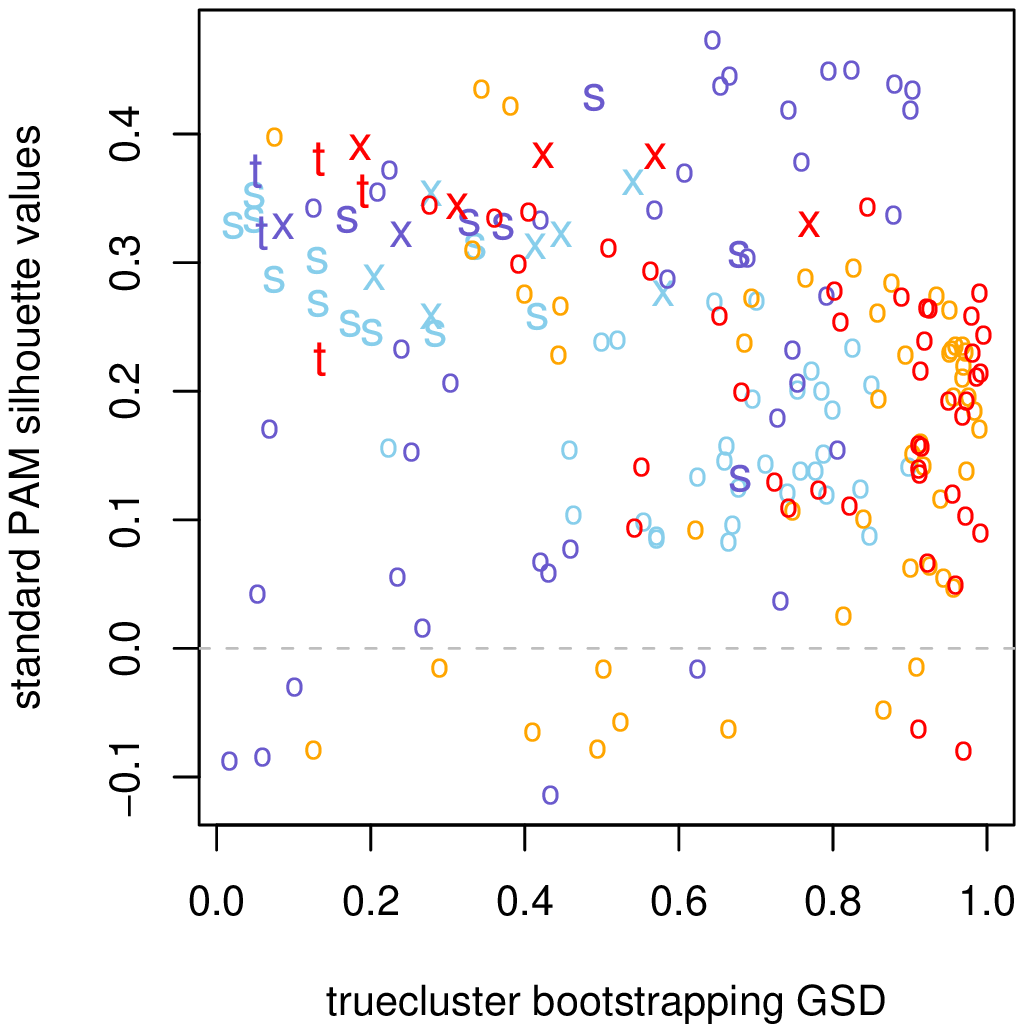}

\caption{Failures to assign true species/gender classes\protect \\
(s = standard PAM fails, t = truecluster fails, x = both fail, o = both ok)\label{cap:Failures-to-assign-species-gender}}
\end{center}
\end{figure}

Figure \ref{cap:Failures-to-assign-species-gender} plots the silhouette values against the truecluster GSDs and shows the localization of the failures with respect to these diagnostics. While standard PAM failures are not associated with low silhouette values, at least all pure truecluster failures have low GSDs, which gives some confidence that GSDs are useful.

\begin{figure}[ht]
\begin{center}
\includegraphics[%
  width=8.6cm,
  angle=-90]{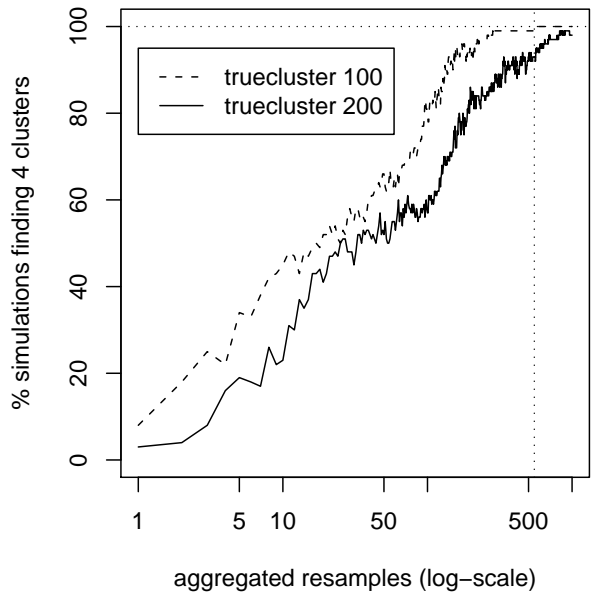}

\caption{Percentage of simulations identifying the 4-cluster solution as best\label{cap:Percentage-of-simulations}}
\end{center}
\end{figure}

To check convergence, we repeated the truecluster bootstrap procedures 100 times and monitored how the $CIC$ results stabilized as we aggregate more and more resamples (Figure \ref{cap:Percentage-of-simulations}). For the truecluster bootstrap procedure, 98\% of the repetitions favoured the 4-cluster solution after aggregating 1000 resamples. The truecluster procedure with subsamples converged faster and reached 100\% decisions for the 4-cluster solutions after aggregating 550 or more resamples.

\begin{figure}[ht]
\begin{center}
\includegraphics[%
  width=12cm,
  keepaspectratio,
  angle=-90]{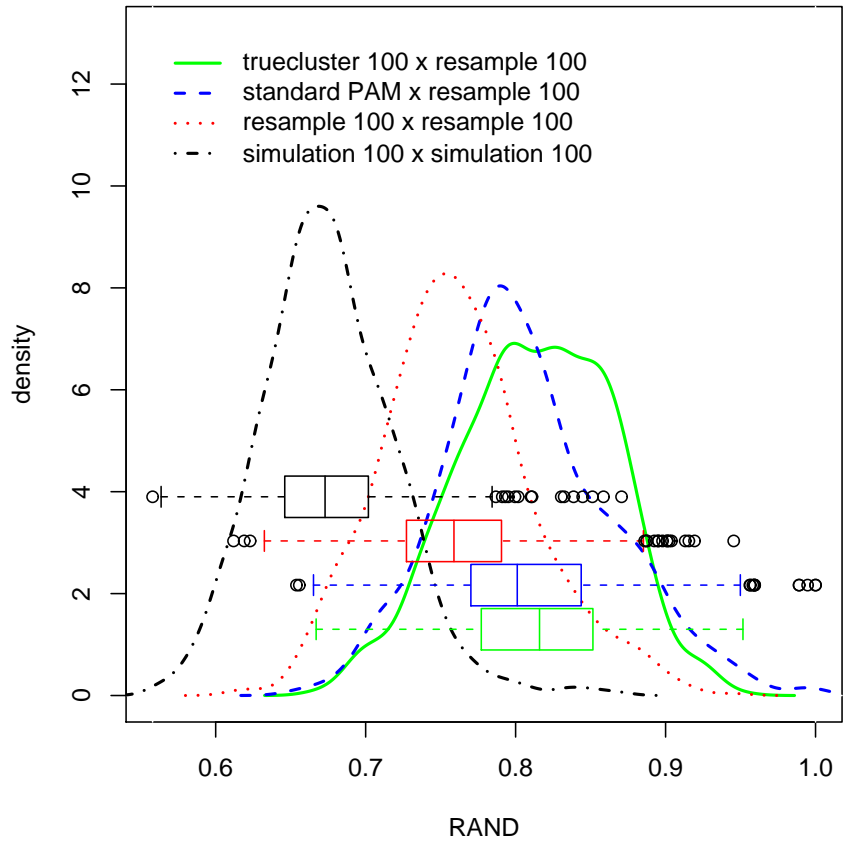}

\caption{Simulation validation and model checking\label{cap:Simulation-validation-and-model-checking}}
\end{center}
\end{figure}

So far we have seen that truecluster was able to identify a plausible best (4-cluster) model and that the estimated cluster memberships agreed quite well with the true classes. Usually in cluster analysis we don't know the true classes: we are looking for them. After having identified a best model in the sample, we still don't know whether the clusters really exist in the population. Resample aggregation does not make sense for the 1-cluster solution, $CIC$ is $0$ by definition. A positive $CIC$ of any $K>1$ solution is an indication of non-randomness of that solution, but we cannot assume that the border between randomness and non-randomness is exactly at $CIC=0$. The $CIC$ is not only a function of the data but also of the resampling scheme, the base cluster algorithm, and the prediction method. In order to check for a non-random pattern using simulation validation, we need the assumption of a reference null distribution. We simulated 1001 successive random samples (n=100) from a multivariate normal distribution with a variance-covariance structure like the original data, then fitted for each a PAM-4 model, and calculated the rand agreement for each of the 1000 successive pairs of cluster solutions (Figure \ref{cap:Simulation-validation-and-model-checking}, black distribution). Similarly, we calculated 1000 rand values from 1001 PAM-4 models on 1001 subsamples (n=100) of the original data (red distribution). Resampling here clearly resulted in higher agreement compared to simulation from a null distribution and the degree of non-overlap between these two distributions is a strong indication that we identified a non-random clustering. 

While the red distribution evaluates agreement between pairs of resamples, the blue distribution in Figure \ref{cap:Simulation-validation-and-model-checking} shows the agreements between resample solutions and the standard sample solution: this is expected to have higher agreements because the standard sample solution is based on the full sample, unlike the solutions from the subsamples which typically include only about 40\% of the sample cases here. This expectation turns out to be true, but note that the green distribution shows even higher agreement between the truecluster solution and the resamples: a clear indication that aggregation of subsamples using only 40\% of the data gives better results than the standard solution based on the full sample. 

In summary, the crab example confirms that it is possible to identify a best model based on aggregating resamples without the need for a reference null distribution, only evaluation of non-randomness requires such an assumption. Truecluster has identified the best (4-cluster) model, the estimated cluster memberships agreed quite well with the true classes, and the case-wise diagnostics have proven useful. Appendix A shows truecluster results for some illustrative artificial datasets. 

\clearpage

\section{Discussion\label{sec:discussion}}

We have presented a statistical framework for robust scalable clustering with model selection for the optimal number of clusters. It assumes that the data has been sampled randomly from an infinite population and mimics this sampling in order to estimate cluster models and to evaluate their quality and stability. Truecluster works with arbitrary definitions of cluster space and clusteriness and, for example, can be applied to symmetry-based k-means \citep{SBKM_SuChou:2001}. The benefits of the truecluster approach are: robust cluster assignments, useful case-wise diagnostics, and a unified framework (also allowing for a unified software interface) to select the best number of clusters. Using subsampling instead of bootstrapping can help to scale `expensive' base algorithms to large samples and subsampling does not automatically reduce solution quality: reducing the resample size turns the base cluster algorithm more towards a `simple base learner' leading to more robust solutions and faster convergence. The computational burden of truecluster may appear high, but it is an economic alternative to expensive manual validation. Instead of a separate evaluation of quality and stability which is computationally expensive as well, truecluster does an integrated evaluation of quality aspects and stability based on generally applicable information-theoretical criteria.

The benefits of truecluster come with two limitations that deserve further research: 1) truecluster gives us a best model for the sample---without relying on the assumption of a reference null distribution---but it does not guarantee that the cluster pattern chosen is non-random and we still might over- or underfit the pattern existing in the population; 2) comparison of $CIC$ across base cluster algorithms requires very cautious interpretation: base cluster algorithms usually differ in model flexibility with respect to the original data, for example, take convex clusters versus arbitrarily shaped clusters. While the model certainty part of the $CIC$ penalizes such greater model flexibility, the model information part of the $CIC$ does not reward it. Consequently, $CIC$ will favour less flexible base cluster algorithms over more flexible ones. This is not a problem as long as we know which kind of model flexibility we need and, therefore, choose an appropriate model class. Model comparisons across models differing in flexibility, to our understanding requires full Bayesian modeling \citep{MacKay:2003} of the original data: solving this generically in software is an enterprise which is much more complicated than the task we have addressed here.

\acks{
We would like to thank Professor Thomas Augustin and Dr. Stefan Pilz for their helpful and encouraging comments on the draft of this paper.
}

\newpage
\appendix
\section*{Appendix A.}
In this appendix we present truecluster results of artificial examples with known cluster structure. The datasets are available online \citep{R:truecluster}.

The colour of each datapoint represents the cluster to which truecluster has assigned it. The reliability of the assignment (generalized silhouette diagnostic) is shown colour-coded in the center of each datapoint. Cases with certain assignment are filled in black, cases with uncertain assignment are filled in white, and cases in between in grey. 

\begin{figure}[ht]
\includegraphics[%
  clip,
  width=1.0\columnwidth,
  angle=-90]{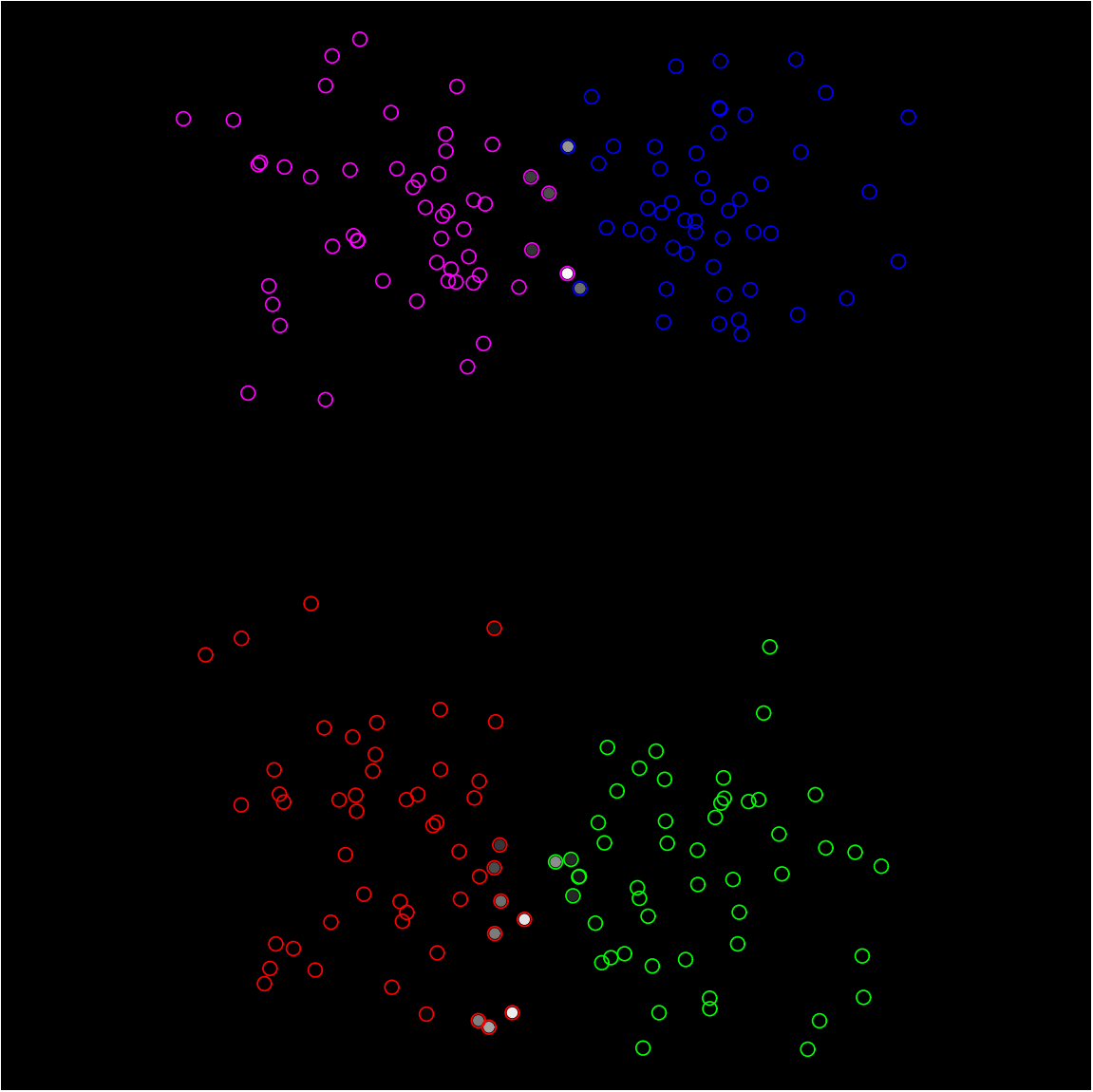}

\caption{Truecluster correctly recognizes the 4 clusters, not only the 2 groups (magenta/blue versus red/green). Cases on the border between magenta/blue or red/green are marked (white) as uncertain assignments. \label{cap:ExampleNoflipper}}
\end{figure}

\begin{figure}[ht]
\includegraphics[%
  clip,
  width=1.0\columnwidth,
  angle=-90]{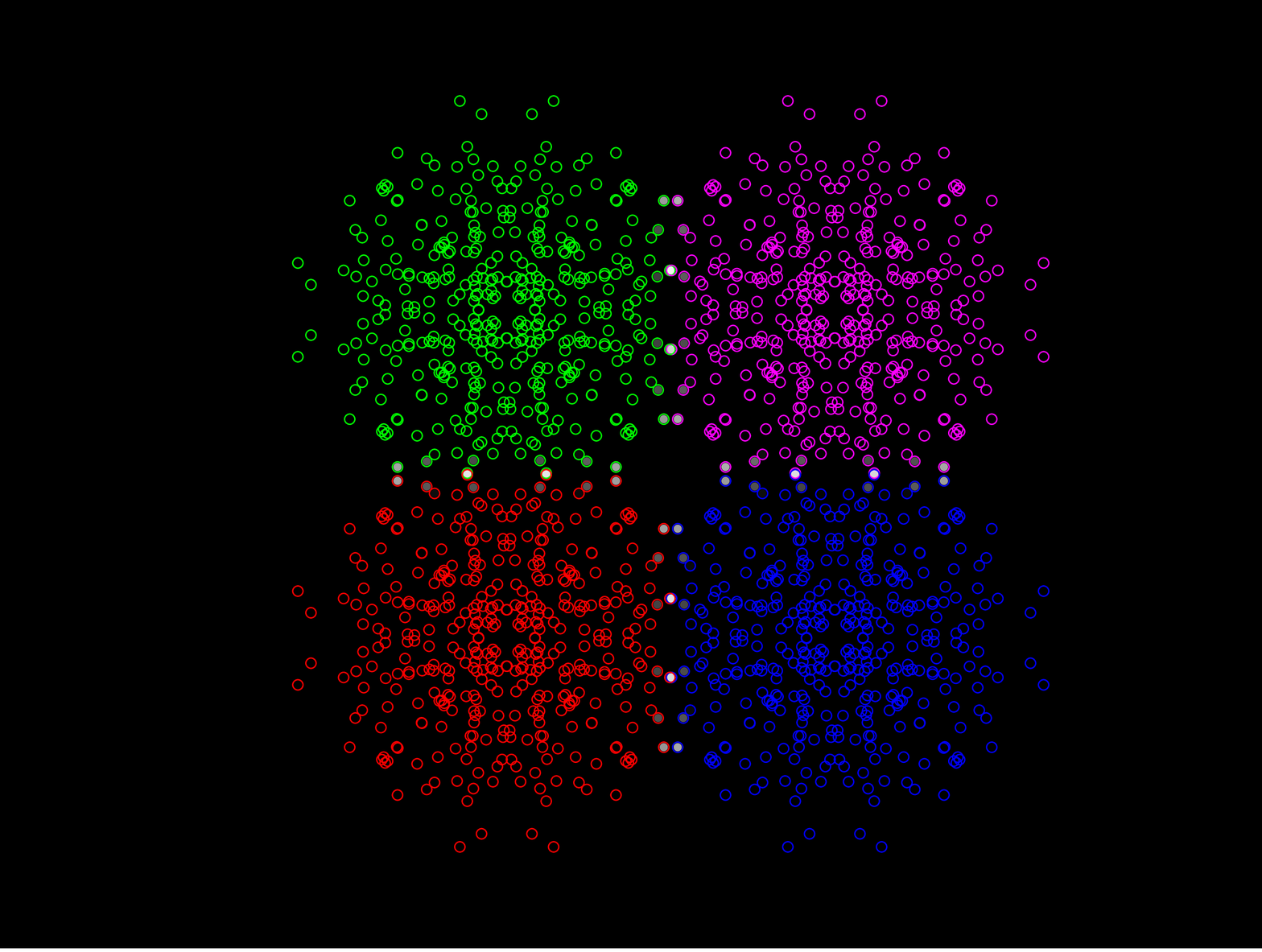}

\caption{Truecluster correctly recognizes the 4 clusters, although they are not well-separated. Cases on the border between the clusters are marked (white) as uncertain assignments. \label{cap:ExampleSymflipper}}
\end{figure}

Figures \ref{cap:ExampleNoflipper} and \ref{cap:ExampleSymflipper}  give examples of convex shaped clusters. The truecluster version of \emph{partitioning around medoids} correctly identifies the correct number of clusters and correctly assigns the cluster memberships.

\begin{figure}[ht]
\includegraphics[%
  clip,
  width=1.0\columnwidth,
  angle=-90]{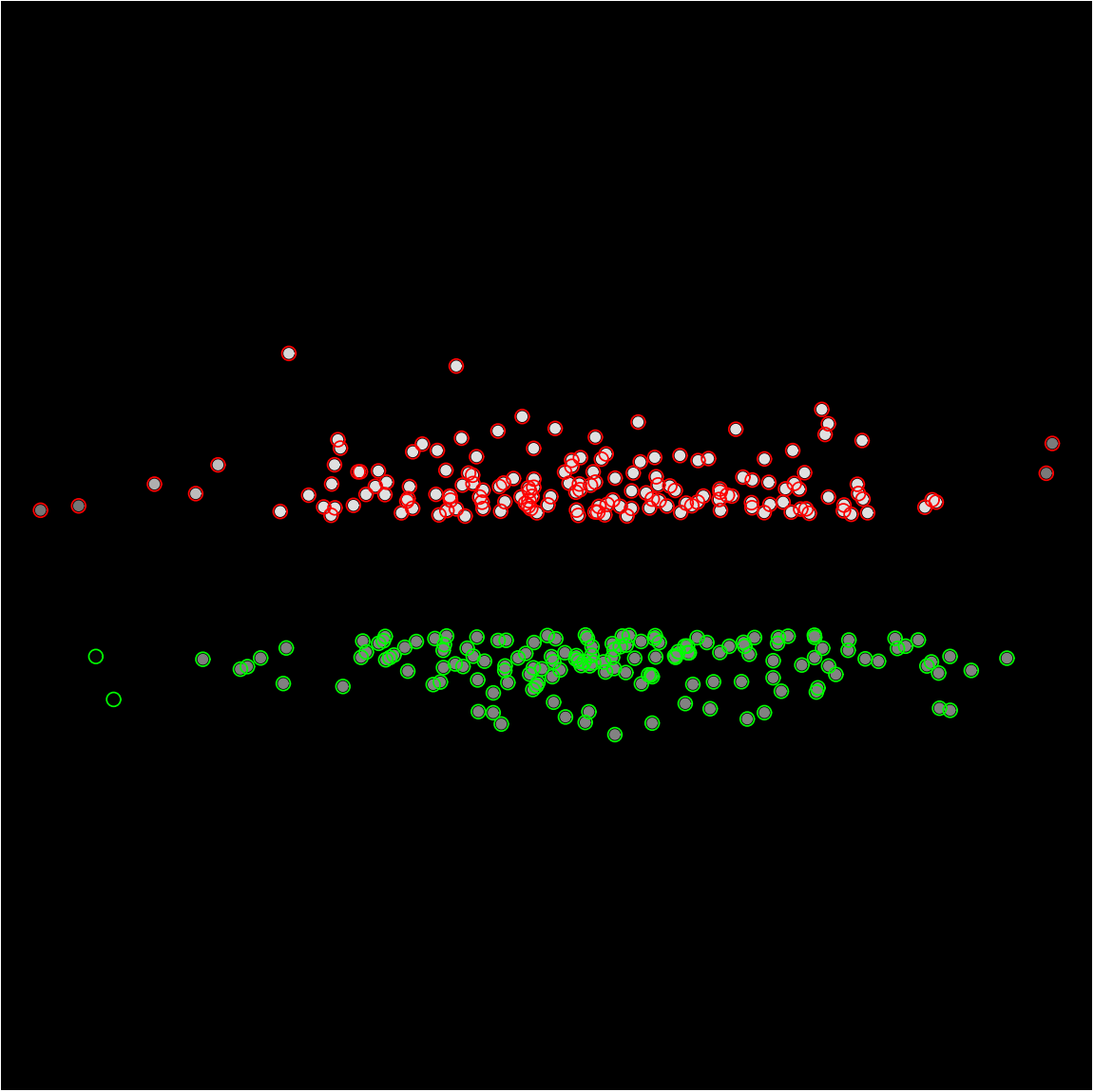}

\caption{Truecluster correctly recognizes 2 elongated clusters \label{cap:ExampleConflict}}
\end{figure}

\begin{figure}[ht]
\includegraphics[%
  clip,
  width=1.0\columnwidth,
  angle=-90]{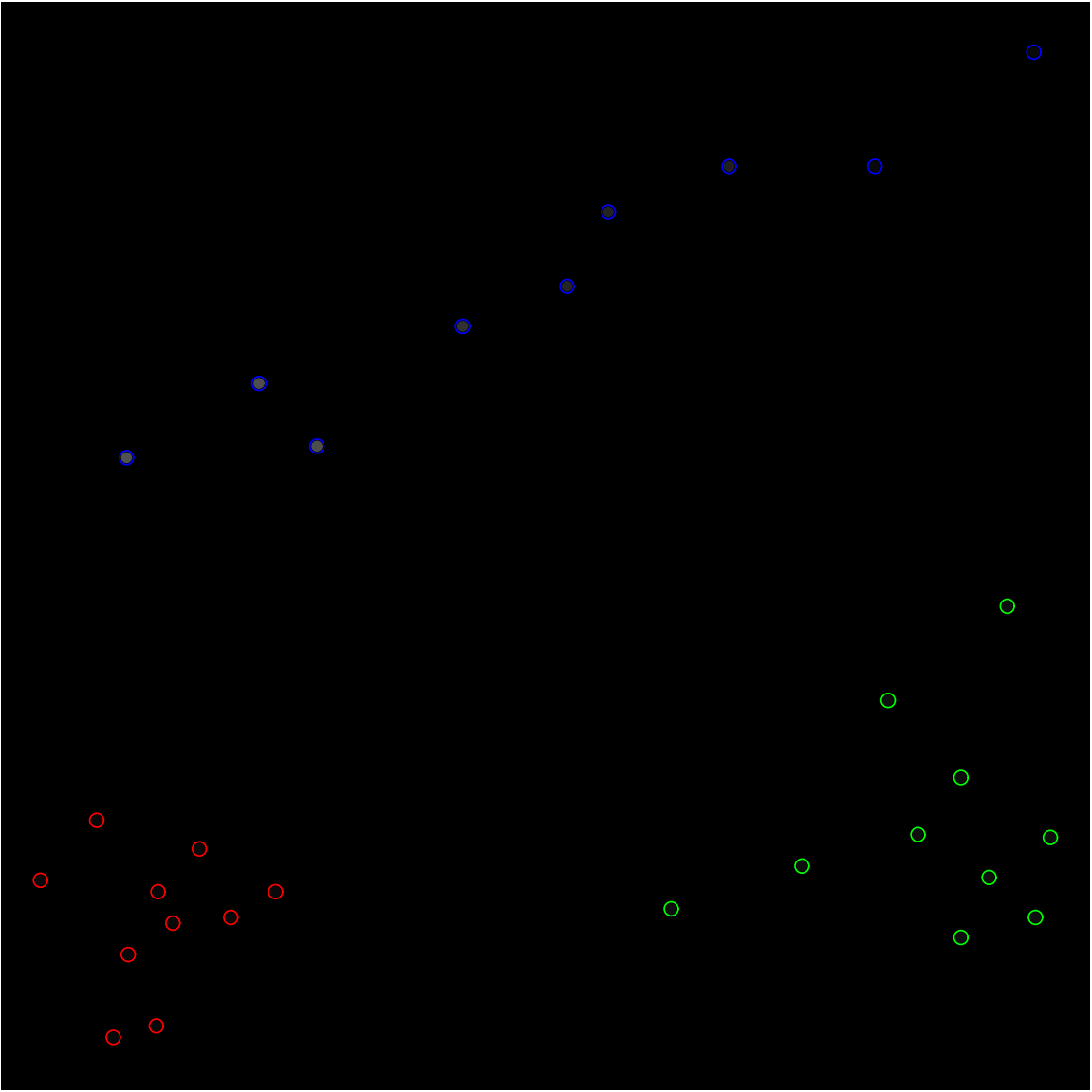}

\caption{Truecluster correctly recognizes the 3 differently shaped clusters.\protect \\
This data is similar to an example from SAS Institute showing that their density-based MODECLUS can detect the 3 clusters \emph{if parameters are chosen correctly}. Truecluster does not require the correct choice of such parameters. \label{cap:ExampleSASModeclus}}
\end{figure}

\begin{figure}[ht]
\includegraphics[%
  clip,
  width=1.0\columnwidth,
  angle=-90]{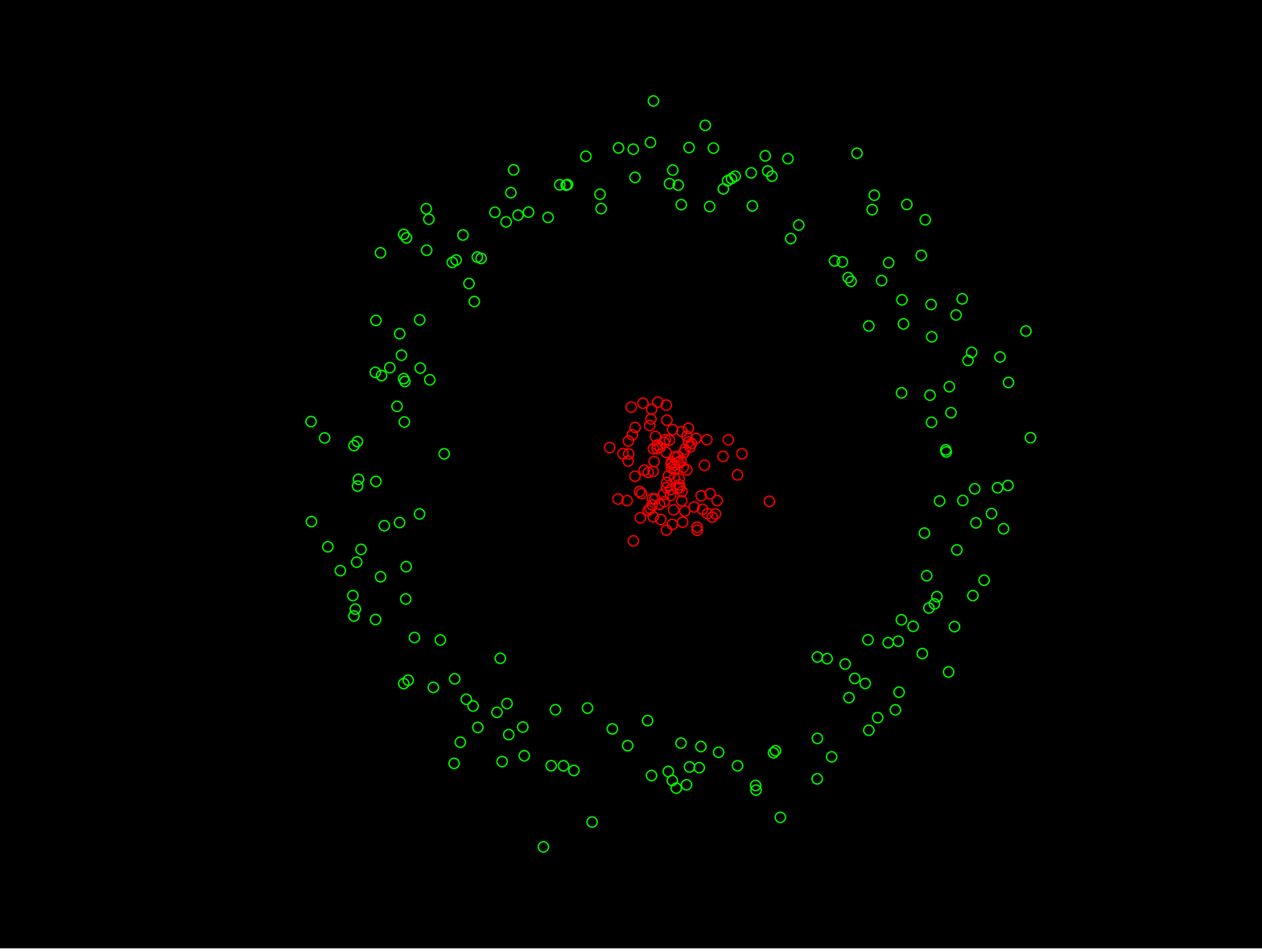}

\caption{Truecluster correctly recognizes 2 clusters, although one is topologically enclosed by the other. \label{cap:ExampleCenterRing}}
\end{figure}

\begin{figure}[ht]
\includegraphics[%
  clip,
  width=1.0\columnwidth,
  angle=-90]{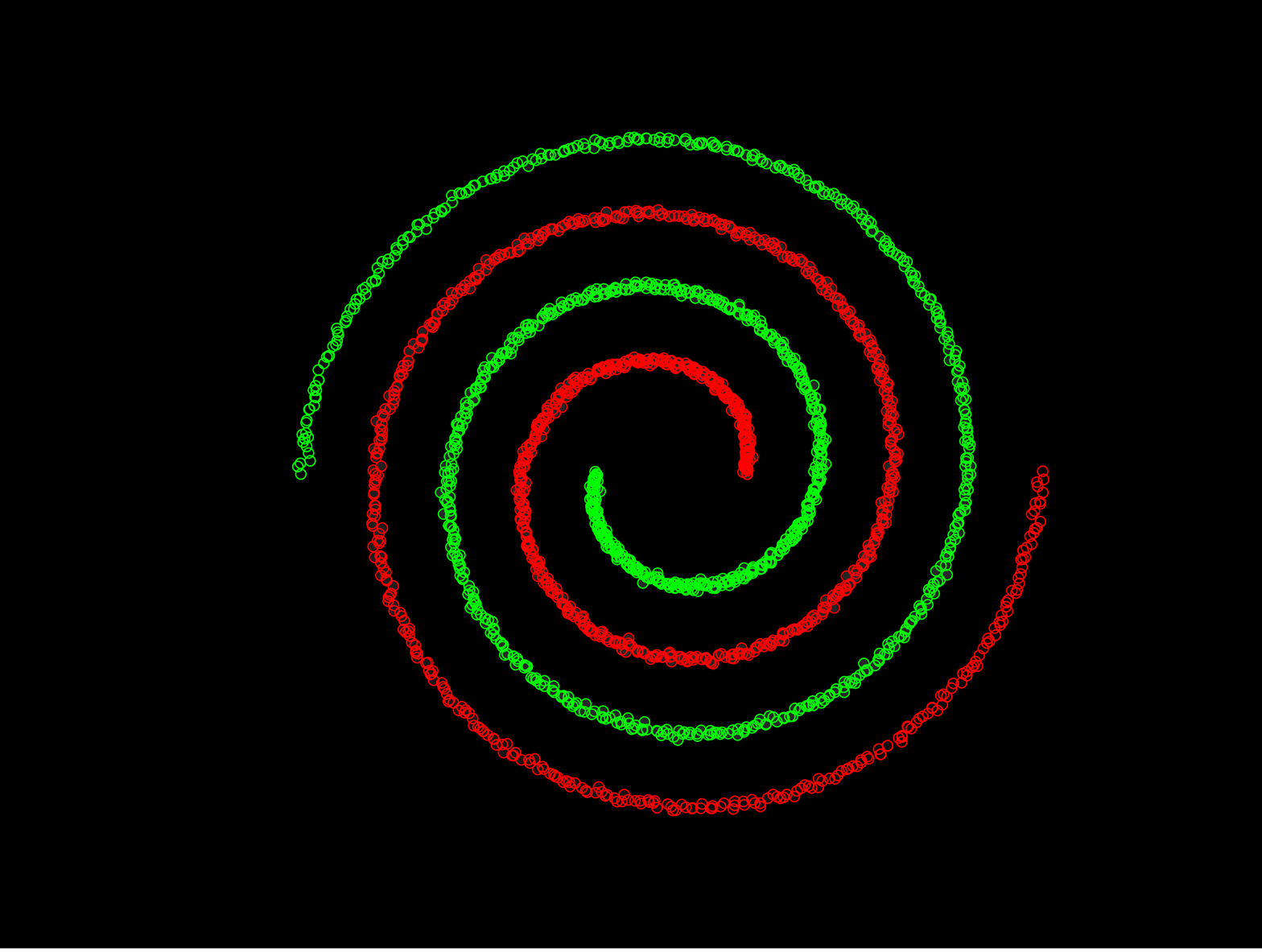}

\caption{Truecluster correctly recognizes 2 spatially complex clusters \label{cap:ExampleSpiral}}
\end{figure}

Figures \ref{cap:ExampleConflict} through \ref{cap:ExampleSpiral} show examples of well-separated arbitrary-shaped clusters. The truecluster version of \emph{single link agglomeration} correctly identifies the correct number of clusters and correctly assigns the cluster memberships.

\newpage
\clearpage

\clearpage

\vskip 0.2in

\bibliography{tc}

\begin{thebibliography}{59}
\providecommand{\natexlab}[1]{#1}
\providecommand{\url}[1]{\texttt{#1}}
\expandafter\ifx\csname urlstyle\endcsname\relax
  \providecommand{\doi}[1]{doi: #1}\else
  \providecommand{\doi}{doi: \begingroup \urlstyle{rm}\Url}\fi

\bibitem[Akaike(1973)]{Akaike:1973}
H.~Akaike.
\newblock Information theory and an extension of the maximum likelihood
  principle.
\newblock In B.N. Petrov and F.~C\'aski, editors, \emph{Second International
  Symposium on Information Theory}, pages 267--281, Budapest, 1973. Akademiai
  Kaid\'o.
\newblock Reprinted in \emph{Breakthroughs in Statistics}, eds Kotz, S. \&
  Johnson, N.L. (1992), volume I, pp. 599--624. New York: Springer.

\bibitem[Akaike(1974)]{Akaike:1974}
H.~Akaike.
\newblock A new look at statistical model identification.
\newblock \emph{IEEE Transactions on Automatic Control}, 19:\penalty0 716--723,
  1974.

\bibitem[Banfield and Raftery(1993)]{MCLUST_BanfieldRaftery:1993}
J.~D. Banfield and A.~E. Raftery.
\newblock Model-based gaussian and non-gaussian clustering.
\newblock \emph{Biometrics}, 49:\penalty0 803--821, 1993.

\bibitem[Bentley(1975)]{Bentley:1975}
John~Louis Bentley.
\newblock Multidimensional binary search trees used for associative searching.
\newblock \emph{Communications of the ACM}, 27\penalty0 (3):\penalty0 180--184,
  1975.

\bibitem[Berkhin(2002)]{berkhin:2002}
Pavel Berkhin.
\newblock Survey of clustering data mining techniques.
\newblock Technical report, Accrue Software, 2002.
\newblock URL \url{citeseer.ist.psu.edu/berkhin02survey.html}.

\bibitem[Bezdek et~al.(1980)Bezdek, Windham, and
  Ehrlich]{BezdekWindhamEhrlich:1980}
James~C. Bezdek, Michael~P. Windham, and Robert Ehrlich.
\newblock Statistical parameters of cluster validity functionals.
\newblock \emph{International Journal of Parallel Programming}, 9:\penalty0
  323--336, 1980.

\bibitem[Bock(1996)]{Bock:1996}
Hans-Herrmann Bock.
\newblock Probability models and hypotheses testing in partitioning cluster
  analysis.
\newblock In P.~Arabie, L.~J. Hubert, and G.~D. Soete, editors,
  \emph{Clustering and Classification.}, pages 377--453. World Scientific
  Publishing, River Edge, NJ, 1996.

\bibitem[Breiman(1996)]{Breiman:1996}
L.~Breiman.
\newblock Bagging predictors.
\newblock \emph{Machine Learning}, 24\penalty0 (2):\penalty0 123--140, 1996.

\bibitem[Breiman(2001)]{Breiman:2001}
L.~Breiman.
\newblock Random forests.
\newblock \emph{Machine Learning}, 45\penalty0 (1):\penalty0 5--32, 2001.

\bibitem[Campbell and Mahon(1974)]{CampbellMahon:1974}
N.~A. Campbell and R.~J. Mahon.
\newblock A multivariate study of variation in two species of rock crab of
  genus leptograpsus.
\newblock \emph{Australian Journal of Zoology}, 22:\penalty0 417--425, 1974.

\bibitem[Cheeseman and Stutz(1996)]{CheesemanStutz:1996}
Peter Cheeseman and John Stutz.
\newblock Bayesian classification (autoclass): Theory and results.
\newblock In U.~Fayyad, G.~Piatetsky-Shapiro, P.~Smyth, and R.~Uthurusamy,
  editors, \emph{Advances in Knowledge Discovery and Data Mining}, pages
  61--83. The AAAI Press, 1996.
\newblock URL \url{citeseer.ist.psu.edu/cheeseman96bayesian.html}.

\bibitem[Chickering and Heckerman(1997)]{ChickeringHeckerman:1997}
David~Maxwell Chickering and David Heckerman.
\newblock Efficient approximations for the marginal likelihood of bayesian
  networks with hidden variables.
\newblock \emph{Machine Learning}, 29\penalty0 (2-3):\penalty0 181--212, 1997.
\newblock URL \url{citeseer.ist.psu.edu/article/chickering96efficient.html}.

\bibitem[Cohen(1960)]{Cohen:1960}
Jacob Cohen.
\newblock A coefficient of agreement for nominal scales.
\newblock \emph{Educational and Psychological Measurement}, 20:\penalty0
  37--46, 1960.

\bibitem[Dempster et~al.(1977)Dempster, Laird, and
  Rubin]{DempsterLairdRubin:1977}
A.~P. Dempster, N.~M. Laird, and D.~B. Rubin.
\newblock Maximum likelihood from incomplete data via the {EM} algorithm (with
  discussion).
\newblock \emph{Journal of the Royal Statistical Society series B},
  39:\penalty0 1--38, 1977.

\bibitem[Dimitriadou et~al.(2002{\natexlab{a}})Dimitriadou, Dolnicar, and
  Weingessel]{DimitriadouDolnicarWeingessel:2002}
E.~Dimitriadou, S.~Dolnicar, and A.~Weingessel.
\newblock An examination of indexes for determining the number of clusters in
  binary data sets.
\newblock \emph{Psychometrika}, 50\penalty0 (2):\penalty0 159--179,
  2002{\natexlab{a}}.

\bibitem[Dimitriadou et~al.(2002{\natexlab{b}})Dimitriadou, Weingessel, and
  Hornik]{DimitriadouWeingesselHornik:2002}
E.~Dimitriadou, A.~Weingessel, and K.~Hornik.
\newblock A combination scheme for fuzzy clustering.
\newblock \emph{Journal of Pattern Recognition and Artificial Intelligence},
  16:\penalty0 901--912, 2002{\natexlab{b}}.

\bibitem[Dolnicar(2003)]{Dolnicar:2003}
S.~Dolnicar.
\newblock Using cluster analysis for market segmentation - typical
  misconceptions, established methodological weaknesses, and some
  recommendations for improvement.
\newblock \emph{Australasian Journal of Market Research}, 11\penalty0
  (2):\penalty0 5--12, 2003.

\bibitem[Dudoit and Fridlyand(2002)]{DudoitFridlyand:2002}
S.~Dudoit and J.~Fridlyand.
\newblock A prediction-based resampling method for estimating the number of
  clusters in a dataset.
\newblock \emph{Genome Biology}, 3\penalty0 (7):\penalty0
  research0036.1--0036.21, 2002.

\bibitem[Dudoit and Fridlyand(2003)]{DudoitFridlyand:2003}
S.~Dudoit and J.~Fridlyand.
\newblock Bagging to improve the accuracy of a clustering procedure.
\newblock \emph{Bioinformatics}, 19\penalty0 (9):\penalty0 1090--1099, 2003.

\bibitem[Efron(1979)]{Efron:1979}
B.~Efron.
\newblock Bootstrap methods: another look at the jacknife.
\newblock \emph{Ann. Statist.}, 7:\penalty0 1--26, 1979.

\bibitem[Efron and Tibshirani(1993)]{EfronTibshirani:1993}
B.~Efron and R.~J. Tibshirani.
\newblock \emph{An Introduction to the Bootstrap}.
\newblock Chapman \& Hall, New York, 1993.

\bibitem[Everitt(1979)]{Everitt:1979}
B.~S. Everitt.
\newblock Unresolved problems in cluster analysis.
\newblock \emph{Biometrics}, 35\penalty0 (1, Perspectives in
  Biometry):\penalty0 169--181, 1979.

\bibitem[Everitt et~al.(2001)Everitt, Landau, and
  Leese]{EverittLandauLeese:2001}
Brian~S. Everitt, Sabine Landau, and Morven Leese.
\newblock \emph{Cluster Analysis}.
\newblock Arnold, London, 2001.

\bibitem[Fraley and Raftery(1998)]{FraleyRaftery:1998}
C.~Fraley and E.~E. Raftery.
\newblock How many clusters? which clustering method? answers via model-based
  cluster analysis.
\newblock \emph{The Computer Journal}, 41\penalty0 (8):\penalty0 578--588,
  1998.

\bibitem[Frey and Dueck(2007)]{FreyDueck:2007}
Brendan~J. Frey and Delbert Dueck.
\newblock Clustering by passing messages between data points.
\newblock \emph{Science}, 315:\penalty0 972--976, 2007.

\bibitem[Gordon(1999)]{Gordon:1999}
A.~D. Gordon.
\newblock \emph{Classification}.
\newblock Chapman \& Hall, Boca Raton, London, New York, Washington, 1999.

\bibitem[Gordon and Vichi(2001)]{GordonVichi:2001}
A.~D. Gordon and M.~Vichi.
\newblock Fuzzy partition models for fitting a set of partitions.
\newblock \emph{Psychometrika}, 66:\penalty0 229--248, 2001.

\bibitem[Gray and Moore(2000)]{GrayMoore:2000}
Alexander~G. Gray and Andrew~W. Moore.
\newblock `n-body' problems in statistical learning.
\newblock Neural Information Processing Systems Conference (NIPS2000), 2000.
\newblock URL \url{books.nips.cc/papers/files/nips13/GrayMoore.pdf}.

\bibitem[Halkidi et~al.(2001)Halkidi, Batistakis, and
  Vazirgiannis]{HalkidiBatistakisVazirgiannis:2001}
M.~Halkidi, Y.~Batistakis, and M.~Vazirgiannis.
\newblock On clustering validation techniques.
\newblock \emph{Intelligent Information Systems Journal}, 17\penalty0
  (2-3):\penalty0 107--145, 2001.

\bibitem[Harrell(2001)]{Harrell:2001}
E.~Jr. Harrell, Frank.
\newblock \emph{Regression Modelling Strategies}.
\newblock Springer, New York, 2001.

\bibitem[Hubert and Arabie(1985)]{HubertArabie:1985}
Lawrence Hubert and Phipps Arabie.
\newblock Comparing partitions.
\newblock \emph{Journal of Classification}, 2:\penalty0 193--218, 1985.

\bibitem[Jain et~al.(2004)Jain, Topchy, Law, and
  Buhmann]{JainTopchyLawBuhmann:2004}
Anil~K. Jain, Alexander Topchy, Martin H.~C. Law, and Joachim~M. Buhmann.
\newblock Landscape of clustering algorithms.
\newblock In \emph{Proc. IAPR International Conference on Pattern Recognition},
  Cambridge, UK, 2004.

\bibitem[Kass and Raftery(1995)]{KassRaftery:1995}
R.~Kass and A.~E. Raftery.
\newblock Bayes factors.
\newblock \emph{Journal of the American Statistical Association}, 90:\penalty0
  773--795, 1995.

\bibitem[Kaufman and Rousseeuw(1990)]{KaufmanRousseeuw:1990}
L.~Kaufman and P.~J. Rousseeuw.
\newblock \emph{Finding Groups in Data: An Introduction to Cluster Analysis.}
\newblock Wiley, New York, 1990.

\bibitem[Krolak-Schwerdt and Eckes(1992)]{Krolak-SchwerdtEckes:1992}
Sabine Krolak-Schwerdt and Thomas Eckes.
\newblock A graph theoretic criterion for determining the number of clusters in
  a data set.
\newblock \emph{Multivariate Behavioral Research}, 27\penalty0 (4):\penalty0
  541--565, 1992.

\bibitem[MacKay(2003)]{MacKay:2003}
David MacKay.
\newblock \emph{Information Theory, Inference, and Learning Algorithms}.
\newblock Cambridge University Press, 2003.

\bibitem[MacQueen(1967)]{KMEANS_MacQueen:1967}
J.~MacQueen.
\newblock Some methods for classification and analysis of multivariate
  observations.
\newblock In L.~Le~Cam and J.~Neymann, editors, \emph{5th Berkley Symp. Math.
  Statist. Prob.}, volume~1, pages 281--297, 1967.

\bibitem[Milligan(1981)]{Milligan:1981}
G.~W. Milligan.
\newblock A monte carlo study of thirty internal criterion measures for cluster
  analysis.
\newblock \emph{Psychometrika}, 46\penalty0 (2):\penalty0 187--199, 1981.

\bibitem[Milligan(1996)]{Milligan:1996}
G.~W. Milligan.
\newblock Clustering validation: results and implications for applied analyses.
\newblock In P.~Arabie, L.~J. Hubert, and G.~D. Soete, editors, \emph{In
  Clustering and Classification.}, pages 341--375. World Scientific Publishing,
  River Edge, NJ, 1996.

\bibitem[Milligan and Cooper(1985)]{MilliganCooper:1985}
G.~W. Milligan and M.~C. Cooper.
\newblock An examination of procedures for determining the number of clusters
  in a data set.
\newblock \emph{Psychometrika}, 50\penalty0 (2):\penalty0 159--179, 1985.

\bibitem[Oehlschl\"agel(2007{\natexlab{a}})]{R:truecluster}
Jens Oehlschl\"agel.
\newblock \emph{Truecluster: an algorithmic framework for robust and scalable
  clustering}, 2007{\natexlab{a}}.
\newblock URL \url{www.truecluster.com}.
\newblock R package version 0.3 (version 1.0 and higher will also be hosted at
  \url{CRAN.R-project.org}).

\bibitem[Oehlschl\"agel(2007{\natexlab{b}})]{Oehlschlaegel:2007b}
Jens Oehlschl\"agel.
\newblock Truecluster matching.
\newblock \emph{submitted to jmlr}, 2007{\natexlab{b}}.

\bibitem[{R Development Core Team}(2003)]{R:W1061}
{R Development Core Team}.
\newblock \emph{R: A language and environment for statistical computing}.
\newblock R Foundation for Statistical Computing, Vienna, Austria, 2003.
\newblock URL \url{www.R-project.org}.
\newblock windows version 1.6.2.

\bibitem[Rand(1971)]{Rand:1971}
W.~M. Rand.
\newblock Objective criteria for the evaluation of clustering methods.
\newblock \emph{Journal of the American Statistical Association}, 66:\penalty0
  846--850, 1971.

\bibitem[Rousseeuw(1987)]{Rousseeuw:1987}
P.~J. Rousseeuw.
\newblock Silhouettes: A graphical aid to the interpretation and validation of
  cluster analysis.
\newblock \emph{J. Comput. Appl. Math.}, 20:\penalty0 53--65, 1987.

\bibitem[Rousseuw et~al.(2004)Rousseuw, Struyf, Hubert, and Hornik]{R:cluster}
Peter Rousseuw, Anja Struyf, Mia Hubert, and Kurt Hornik.
\newblock \emph{cluster: Functions for clustering (by Rousseeuw et al.)}, 2004.
\newblock S original by Peter Rousseuw and Anja Struyf and Mia Hubert. R port
  by Kurt Hornik, R package version 1.6-4.

\bibitem[Sarle(1983)]{Sarle:1983}
W.~Sarle.
\newblock Cubic clustering criterion.
\newblock Technical Report A-108, SAS Institute, Inc., 1983.

\bibitem[Schwarz(1978)]{Schwarz:1978}
G.~Schwarz.
\newblock Estimating the dimension of a model.
\newblock \emph{Annals of Statistics}, 6:\penalty0 461--464, 1978.

\bibitem[Shannon(1948)]{Shannon:1948}
C.~E. Shannon.
\newblock A mathematical theory of communication.
\newblock \emph{Bell System Tech. J.}, 27:\penalty0 379--423, 1948.
\newblock as reprinted in ``The Mathematical Theory of Communication'', C. E.
  Shannon and W. Weaver, University of Illinois Press, Champaign­Urbana (1963).

\bibitem[Smyth(1996)]{Smyth:1996}
P.~Smyth.
\newblock Clustering using monte carlo cross-validation.
\newblock In \emph{Proceedings of the 2nd International Conference on Knowledge
  Discovery and Data Mining}, pages 126--133. AAAI Press, 1996.

\bibitem[Strehl and Ghosh(2002)]{StrehlGhosh:2002}
A.~Strehl and J.~Ghosh.
\newblock Cluster ensembles --- a knowledge reuse framework for combining
  multiple partitions.
\newblock \emph{Journal of Machine Learning Research}, 3:\penalty0 583--617,
  2002.

\bibitem[Su and Chou(2001)]{SBKM_SuChou:2001}
M.~C. Su and C.~H. Chou.
\newblock A modified version of the {K-Means} algorithm with a distance based
  on cluster symmetry.
\newblock In \emph{IEEE Trans. on Pattern Analysis and Machine Intelligence},
  volume 23(6), pages 674--680, June 2001.

\bibitem[Thorndike(1953)]{Thorndike:1953}
R.~L. Thorndike.
\newblock Who belongs in the family?
\newblock \emph{Psychometrika}, 4:\penalty0 267--276, 1953.

\bibitem[Tibshirani et~al.(2001{\natexlab{a}})Tibshirani, Walther, and
  Hastie]{TibshiraniWaltherHastie:2001}
R.~Tibshirani, G.~Walther, and T.~Hastie.
\newblock Estimating the number of clusters in a dataset via the gap statistic.
\newblock \emph{J. Royal Statist. Soc. B.}, 63:\penalty0 411--424,
  2001{\natexlab{a}}.

\bibitem[Tibshirani et~al.(2001{\natexlab{b}})Tibshirani, Walther, Botstein,
  and Brown]{TibshiraniWaltherBotsteinBrown:2001}
Robert Tibshirani, Guenther Walther, David Botstein, and Patrick Brown.
\newblock Cluster validation by prediction strength.
\newblock Technical report, Stanford University, 2001{\natexlab{b}}.

\bibitem[Venables and Ripley(1994)]{VenablesRipley:1994}
W.~N. Venables and B.~D. Ripley.
\newblock \emph{Modern Applied Statistics with S-PLUS}.
\newblock Statistics and Computing. Springer, New York, Berlin, 3rd edition,
  1994.

\bibitem[Venables and Ripley(2002)]{R:MASS}
W.~N. Venables and B.~D. Ripley.
\newblock \emph{MASS Modern Applied Statistics with S-PLUS}, 2002.
\newblock R package version 7.0-10.

\bibitem[Wong and Schaack(1982)]{WongSchaack:1982}
M.~A. Wong and C.~Schaack.
\newblock Using the k-th nearest neighbor clustering procedure to determine the
  number of subpopulations.
\newblock In \emph{Proceedings of the Statistical Computing Section}, pages
  40--48, 1982.

\bibitem[Zaiane et~al.(2002)Zaiane, Foss, Lee, and
  Wang]{ZaianeFossLeeWang:2002}
O.~R. Zaiane, A.~Foss, C.~H. Lee, and W.~Wang.
\newblock On data clustering analysis: Scalability, constraints, and
  validation.
\newblock In \emph{Proceedings of the 6th Pacific-Asia Conference on Knowledge
  Discovery and Data Mining (PAKDD)}, pages 28--39, 2002.

\end{thebibliography}

\end{document}